\documentclass[journal,transmag]{IEEEtran}

\ifCLASSINFOpdf
\else
\fi
\usepackage{cite}
\usepackage{graphicx}
\usepackage{bm}
\usepackage{multirow} 
\usepackage{booktabs}
\usepackage{amssymb}
\usepackage{amsmath}
\usepackage{tabularx}
\usepackage{algorithm} 
\usepackage{algorithmic}
\usepackage{graphicx}
\usepackage{subfigure}

\begin{document}
	%
	\title{Effective Multi-view Registration of Point Sets based on \\Student’s T Mixture Model}

	
	\author{\IEEEauthorblockN{
	        Yanlin Ma\IEEEauthorrefmark{1},
			Jihua Zhu\IEEEauthorrefmark{1}, 
			Zhongyu Li\IEEEauthorrefmark{1},
			Zhiqiang Tian\IEEEauthorrefmark{1},
		    Yaochen Li\IEEEauthorrefmark{1},}
		\IEEEauthorblockA{\IEEEauthorrefmark{1}School of Software Engineering, Xi'an Jiaotong University, Xi’an 710049, China}
		\thanks{Corresponding author: Jihua Zhu (email: zhujh@xjtu.edu.cn).}}

	\markboth{Journal of \LaTeX\ Class Files,~Vol.~14, No.~8, August~2015}%
	{Shell \MakeLowercase{\textit{et al.}}: Bare Demo of IEEEtran.cls for IEEE Transactions on Magnetics Journals}
	%



	\IEEEtitleabstractindextext{%
		\begin{abstract}
			Recently, Expectation-maximization (EM) algorithm has been introduced as an effective means to solve multi-view registration problem. Most of the previous methods assume that each data point is drawn from the Gaussian Mixture Model (GMM), which is difficult to deal with the noise with heavy-tail or outliers. Accordingly, this paper proposed an effective registration method based on Student’s t Mixture Model (StMM). More specially, we assume that each data point is drawn from one unique StMM, where its nearest neighbors (NNs) in other point sets are regarded as the t-distribution centroids with equal covariances, membership probabilities, and fixed degrees of freedom. Based on this assumption, the multi-view registration problem is formulated into the maximization of the likelihood function including all rigid transformations. Subsequently, the EM algorithm is utilized to optimize rigid transformations as well as the only t-distribution covariance for multi-view registration. Since only a few model parameters require to be optimized, the proposed method is more likely to obtain the desired registration results. Besides, all t-distribution centroids can be obtained by the NN search method, it is very efficient to achieve multi-view registration. What's more, the t-distribution takes the noise with heavy-tail into consideration, which makes the proposed method be inherently robust to noises and outliers. Experimental results tested on benchmark data sets illustrate its superior performance on robustness and accuracy over state-of-the-art methods.
		\end{abstract}
		
		}

	\maketitle

	\IEEEdisplaynontitleabstractindextext

	%
	\IEEEpeerreviewmaketitle

	\section{Introduction}
	%
	%
	%
	%
	\IEEEPARstart{P}{oint} set registration is a fundamental and important problem in many domains, such as computer vision \cite{yang2015go, lei2017fast,zhao20183}, robotics \cite{jiang2019simultaneous,yu2015semantic} and computer graphics \cite{roy2004wide,aiger20084,dai2017bundlefusion}. With the development of scanning devices, it becomes very easy to acquire 3D scan data from the object or scene. Due to the limited view or occlusion, it is difficult to obtain the scanning data of the whole object or scene at one time. To reconstruct the 3D model, it is necessary to acquire multiple point sets at different view points and then unify these point sets into the same coordinate frame, which leads to the multi-view registration problem.
	
	For the point set registration, Iteration closest point (ICP) \cite{besl1992method,chen1992object} is one of the most popular methods. Given initial rigid transformation, it alternatively implements the establishment of point correspondence and the optimization of rigid transformation to achieve pair-wise registration with good efficiency. Since it is difficult to find the real point correspondence between two point sets being registered, some mixture probabilistic models, such as Gaussian Mixture Model (GMM) \cite{myronenko2010point,jian2010robust,gao2019filterreg} and Student’s t Mixture Model (StMM) \cite{peel2000robust} were proposed to improve the registration accuracy. These probabilistic methods are effective for pair-wise registration, they are further extend to solve multi-view registration problem \cite{evangelidis2017joint}. Most of these methods assume that all data points are realizations of one central GMM or StMM. Since the registration performance is determined by the number of mixture components, there is a trade of between accuracy and efficiency. Besides, GMM-based methods suffer from the weakness of GMM in the noise with heavy-tail and are difficult to eliminate the influence of outliers in registration. Accordingly, it is difficult to obtain very accurate registration results. What's more, these methods require to estimate massive model parameters, which make it easy to be trapped into local minimum.
	
	To address these problems, we propose a novel method of multi-view registration based the Student’s t Mixture Model (StMM) \cite{peel2000robust}. Usually, only one data point may be acquired from one object or scene position in each point set. For the registration, it only requires to find one correspondence of each data point from each other point sets. Due to the noises, it is difficult to establish the accurate point correspondences between different point sets. Accordingly, it is reasonable to suppose that one data point is drawn from the t-distribution, whose centroid is attached to the corresponding point in other point set. To utilize all available information, we further suppose that each data point is drawn from one unique StMM, where its NNs in other point sets are regarded as the t-distribution centroids with equal covariances, membership probabilities, and the fixed degrees of freedom.
    
    In the proposed method, the number of t-distribution components in StMM is automatically determined by the number of scans being registered. Therefore, it is no trade-off between registration accuracy and efficiency. Since the t-distribution contains more heavy tail than that of Gaussian distribution, StMM can model scan data without any prior knowledge of the degree of outliers and noises. Accordingly, it is easy to obtain very accurate registration results. What's more, all the t-distribution centroids are efficiently determined by the NN search, so it only requires to estimate one covariance as well as some rigid transformations, which make it more likely to obtain desired registration results. Experimental results test on six bench mark data sets will illustrate its superior performance for the multi-view registration.

	The remainder of this paper is organized as follows: Section 2  sketchily discusses the related work. Section 3 formulates the multi-view registration problem by StMM. Section 4 optimizes rigid transformations and other model parameters under the framework of expectation maximization. Section 5 tests the proposed method on six bench mark data sets and compares it with three state-of-the-art methods. Finally, some conclusions are drawn in Section 6.

	\section{Related Work}
	This section mainly introduces related works on point set registration, which can be generally divided into two categories: pair-wise registration and multi-view registration.
	
	\subsection{Pair-wise registration}
	
	For the study of pair-wise registration, it can be categorized two sub-problems, i.e., precise registration and coarse registration. ICP \cite{besl1992method,chen1992object} is one of the classic methods for precise registration proposed by P. J. Besl et al. This approach takes the least squares estimator as the objective function, which is easily biased by outliers and cannot handle non-overlapped point clouds. Then, Chetverikov et al. \cite{chetverikov2005robust} proposed the trimmed iterative closest point algorithm (TrICP), which alleviates the limitations of ICP by introducing the overlap percentage to identify the registration area. However, it also suffers from the same weakness as the least squares estimator of ICP. To this end, GMM had been used in pair-wise registration, such as CPD \cite{myronenko2010point}, GMMReg \cite{jian2010robust} and FilterReg \cite{gao2019filterreg}. CPD \cite{myronenko2010point} assumed that the data point set introduced a GMM controlled by transformation. Conversely, FilterReg \cite{gao2019filterreg} assumed that observed point set introduced a GMM. Whereas, GMMReg \cite{jian2010robust} considered that both point sets introduced GMM. Then, the three methods employed the EM algorithm in the optimization process. Instead of least squares estimation, probability model can robust to noise and outliers, significantly outperforming ICP and TrICP. However, GMM cannot achieve registration results, when the point set contains noise with heavy-tail. Zhou et al. \cite{zhou2014robust} proposed a more robust method through StMM \cite{peel2000robust}. They treat observed point set as StMM centroids, and then fit the StMM centroids to the data point set. T-distribution has heavier tails than that of Gaussian distribution, so this method can achieve more robust and accurate registration results than GMM when large amounts of noise with the heavy-tail are existed. Whereas, it is inevitable to fall into the local optimum, if the registration starts with a terrible initialization. 
	
	Coarse registration can provide a rough initial rigid transformation for precise registration, where it mainly uses feature extraction and randomized strategies. The 4PCS \cite{aiger20084} works on the RANSAC \cite{fischler1981random} framework, which constructs sampled tuples of four co-planar points, with the employment of the LCP (Largest Common Pointset) strategy to search the four co-planar points with the maximum overlap to obtain the registration result. To improve the efficiency of 4PCS and reduce the selection of invalid point pairs, Super4PCS \cite{mellado2014super} was proposed. What's more, R. B. Rusu et al. \cite{rusu2008aligning,rusu2008persistent} proposed the point feature histogram (PFH) algorithm, which forms a multi-dimensional histogram to describe the neighbors of a point geometrically by spatial difference. The information provided by the histogram is invariant to translation and rotation with robustness. To reduce the computational times of PFH, they proposed fast point feature histogram (FPFH) \cite{rusu2009fast}. More recently, Huu M. Le et al. \cite{le2019sdrsac} proposed the SDRSAC algorithm, which combined randomized techniques and graph matching as an effective random sampling to register solving by tight semidefinite (SDP) relaxation.

	\subsection{Multi-view registration}
	
	Multi-view registration can simultaneously register multiple point sets. Intuitively, Sequential pair-wise registration strategy is used for multi-view registration \cite{chen1992object}. However, the concatenation of them along with a cycle cannot achieve satisfactory results, even if each pair-wise registration is correct. Further, if the results of pair-wise registration contain incorrect correspondences, the situation is worse inevitably. To equally distribute the registration errors, Bergevin et al. \cite{bergevin1996towards} proposed a multi-view registration algorithm which considers all views as a whole, where this approach defines initial transformations as a star-shaped topology network. Then, each view is sequentially set as the center of the network to iteratively compute the incremental transformation matrices with remaining views. Subsequently, Guo et al. \cite{guo2014accurate} proposed a shape growing multi-view registration algorithm. It selects a point set as the seed point set, and then iteratively grows by performing pair-wise registration between itself and the remaining point sets, where the pair-wise registration adopts RoPS \cite{guo2013rotational} feature extraction combined with the coarse-to-fine based ICP algorithm to improve the accuracy. Although these methods optimize registration accuracy, the sequential registration problems are still not solved.
	
	To this end, Williams et al. \cite{williams2001simultaneous} proposed a method to recover global optimal transformation by minimizing across all view correspondences simultaneously. However, it is time-consuming due to the construction of massive correspondences. Meanwhile, Huber et al. \cite{huber2003fully} applied spanning tree based methods to the multi-view registration, where pair-wise registrations construct a model graph to build a spanning tree for multi-view registration. Since then, some other methods using spanning tree have been proposed \cite{tombari2010unique,bariya20123d}. In addition, if objects are symmetric, the transformation of two views could be incorrect in global transformation, even if they are registered correctly in pair-wise registration. X. Mateo et al. \cite{mateo2014bayesian} introduced Bayesian frame to utilize different weights which represented the reliability of the correspondences between different views, with the purpose of detecting the possibility of some incorrect registrations. Thereby, it minimizes incorrect pair-wise registration impact in the global as far as possible. Since this method requires to estimate massive variables, its efficiency is worse. These methods avoid sequential registration, but the total registration errors are still not alleviated.    
	
	Therefore, Govindu and Pooja \cite{govindu2013averaging} proposed a motion averaging algorithm, which can estimate the global optimal registration based on a group of pair-wise registration. However, this approach requires a lot of reliable pair-wise transformation, otherwise the result is worse. Considering the different reliability of each pair-wise registration, Guo et al. \cite{guo2018weighted} proposed the weighted motion averaging algorithm. Arrigoni et al. \cite{arrigoni2016global,arrigoni2018robust} proposed the low-rank and sparse (LRS) matrix decomposition algorithm. Through the matrix decomposition, this algorithm can recover global transformation from a block matrix, where the block matrix contains available pair-wise registration. To pay more attentions on reliable pair-wise registrations, Jin et al. \cite{jin2018multi} proposed the weighted LRS algorithm, which could achieve more accurate and robust registration results than original methods.
	
    Recently, multi-view registration methods through GMM have been developed. Evangelidis et al. \cite{evangelidis2017joint} proposed JRMPC, which assumes that the point sets have $K $ Gaussian centroids by clustering to generate each data point. In this way, the global information can be combined for registration to avoid the error accumulation problem. The GMM can register with strong robustness in small quantity of noises. However, it may dramatically fluctuate in noise with heavy-tail. Hence, GMM needs to add a weighted uniform distribution component, which enhances robustness but reduces accuracy. Furthermore, JRMPC requires to estimate massive parameters, which makes it easy to be trapped into local optimum. To this end, Ravikumar et al. \cite{ravikumar2018group} introduced StMM to solve a part shortcoming of JRMPC. T-distribution is a family of alternative distribution which has heavier tails than that of Gaussian distribution \cite{peel2000robust}. Therefore, it can robustly register noise samples with the heavy-tail. As previously mentioned, these methods suffer from accuracy losses by clustering. The more centroids are set, the more accurate the registration result can be achieved, while the higher computational complexity existed. In a word, they need to consider the trade-off between efficiency and accuracy. Zhu et al. \cite{zhu2020registration} proposed the EMPMR algorithm, which increased the efficiency by making the number of GMM centroids reduce to the number of point sets. Although EMPMR can dramatically reduce the computational complexity, the weakness of GMM still need to be solved. Here, we propose a novel multi-view registration method using StMM with expectation maximization, by considering these unresolved issues.

	\section{Problem Formulation}
	\subsection{T-distribution}
	We let $x = \left[ {{x_1},{x_2}, \ldots ,{x_n}} \right]$ denote a $d$-dimensional sample. Then, we assume that each data point subjects to a $d$-dimensional Gaussian distribution and the probability density function is defined as:
	\begin{equation}
	    {{f}_{N}}\left( x;\mu ,\Sigma  \right)=\frac{1}{{{\left( 2\pi  \right)}^{\frac{2}{d}}}{{\left| \Sigma  \right|}^{\frac{1}{2}}}}{{e}^{-\frac{1}{2}{{\left( x-\mu  \right)}^{T}}{{\Sigma }^{-1}}\left( x-\mu  \right)}},
	    \label{eq_1}
	\end{equation}
	where $\mu $ and $\Sigma $ indicate the mean and covariance matrix of Gaussian distribution, respectively.
	
	Due to the short tail, the Gaussian distribution is sensitive to noises and outliers. While, the t-distribution with the same mean $\mu $ as Gaussians can overcome this problem. T-distribution has heavier tail than that of Gaussian, which means t-distribution has a larger covariance matrix to exclude the interference from noises as well as outliers and preserve the characteristics of the data. The derivation of t-distribution is imposing a Gamma distribution $U\sim gamma\left( {v}/{2}\;,{v}/{2}\; \right)$ as a prior on the covariance $\Sigma $ of the Gaussian distribution and then marginalizing out the scaling weights $u$ as follows:
	\begin{equation}
	    {{f}_{T}}\left( x;\mu ,\Sigma ,v \right)=\int{{{f}_{N}}\left( x;\mu ,{\Sigma }/{u}\; \right)}\mathcal{G}\left( u;{v}/{2}\;,{v}/{2}\; \right)du.
	    \label{eq_2}
	\end{equation}
Here, the probability density function of $\mathcal{G}\left( u;\alpha ,\beta  \right)$ is defined by:
	\begin{equation}
	    {{f}_{G}}\left( u;\alpha ,\beta  \right)=\frac{{{\beta }^{\alpha }}{{u}^{\alpha -1}}}{\Gamma \left( \alpha  \right)}{{e}^{-\beta u{{I}_{\left( 0,\infty  \right)}}\left( u \right)}},
	    \label{eq_3}
	\end{equation}
where, ${{I}_{\left( 0,\infty  \right)}}\left( u \right)$ is the indicator function, ${{I}_{\left( 0,\infty  \right)}}\left( u \right)=1$ for $u>0$ and ${{I}_{\left( 0,\infty  \right)}}\left( u \right)=0$ elsewhere. We can then obtain the probability density function of t-distribution (Eq.~(\ref{eq_4})) by evaluating the integral show in Eq.~(\ref{eq_2}):
	\begin{equation}
	    {{f}_{T}}\left( x;\mu ,\Sigma ,v \right)=\frac{\Gamma \left( \frac{v+d}{2} \right)}{{{\left| \Sigma  \right|}^{\frac{1}{2}}}\Gamma \left( \frac{v}{2} \right){{\left( \pi v \right)}^{\frac{d}{2}}}{{\left[ 1+\frac{{{\left( x-\mu  \right)}^{T}}{{\Sigma }^{-1}}\left( x-\mu  \right)}{v} \right]}^{\frac{v+d}{2}}}},
	    \label{eq_4}
	\end{equation}
	where $\Gamma \left( \cdot  \right)$ denotes the gamma function.
		\begin{figure}
		\centering
		\includegraphics[width=0.75\columnwidth]{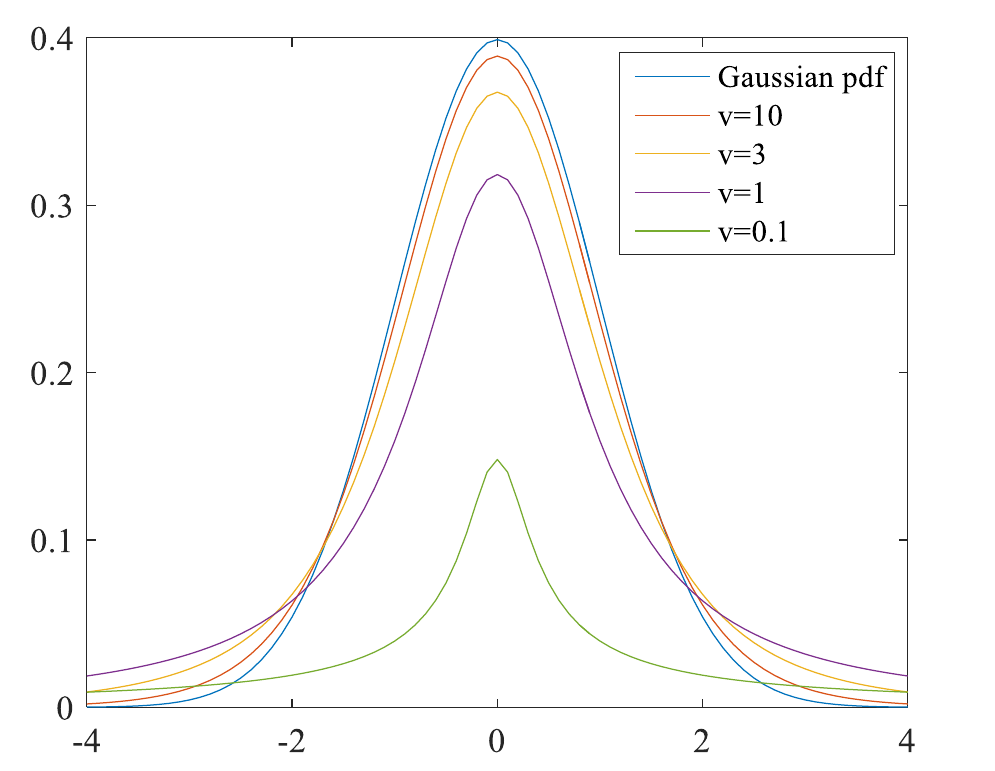}
		\caption{The probability density function of the Gaussian distribution and t-distribution with the different parameter $v$.}
		\label{fig_1}
	\end{figure}
	
	The parameter $v$ is the degree of freedom that control the tail of t-distribution. If $v>1$, $\mu $ is the same as Gaussian mean. If $v>2$, $v{{\left( v-2 \right)}^{-1}}\Sigma $ denotes the covariance matrix of t-distribution, where $\Sigma $ denotes the covariance of corresponding Gaussian distribution. As $v\to \infty $, the t-distribution will be equivalent to the Gaussian distribution. Both above content and Fig.~\ref{fig_1} articulate that the family of t-distribution provides an alternative distribution which has heavier tails than that Gaussian distribution.
	
	\subsection{Registration of multi-view point sets using Student’s t Mixture Model}
\begin{figure}
		\centering
		\includegraphics[width=1.0\columnwidth]{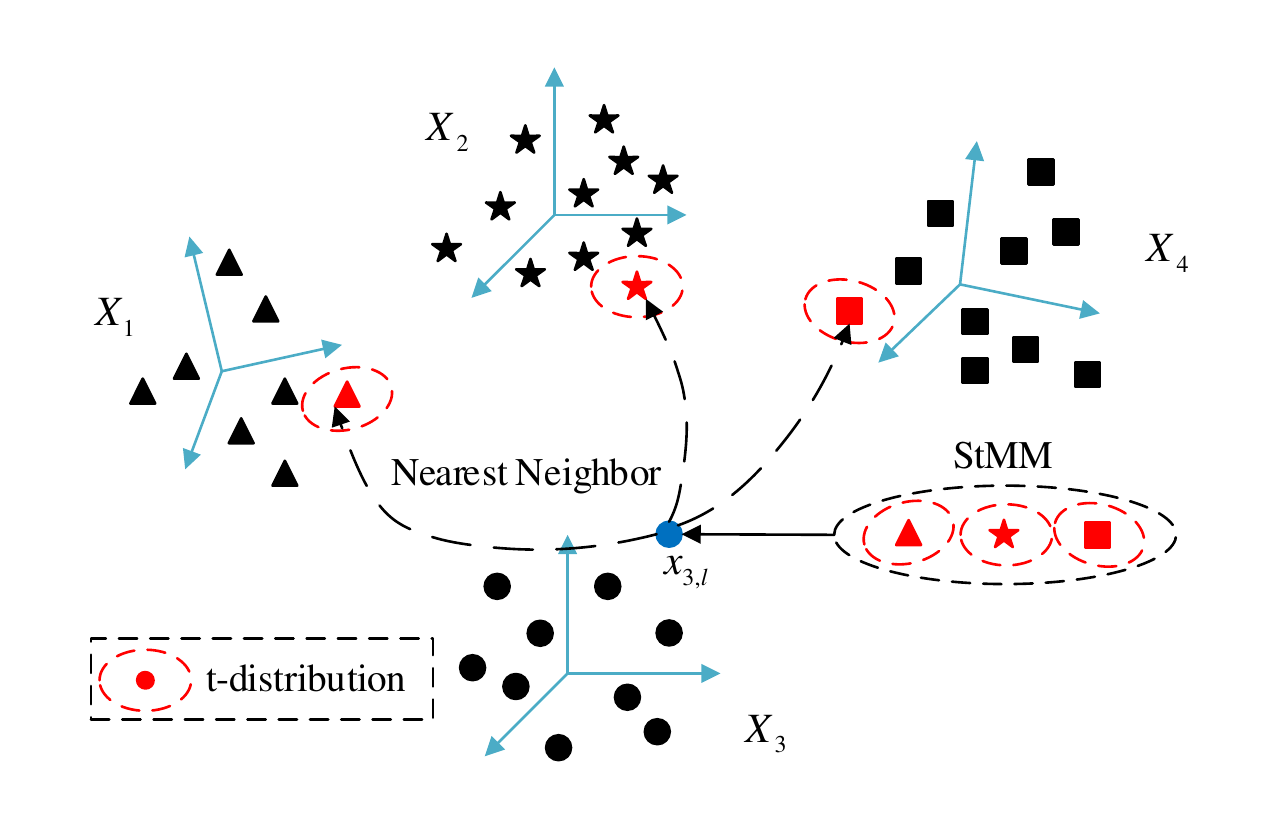}
		\caption{Illustration of proposed of registration method based on StMM. It assumes that each data point ${{x}_{i,l}}$ is generated from one unique StMM, where its nearest neighbors in other point sets are regarded as the t-distribution centroids with equal covariances, membership probabilities and the fixed degree of freedom.}
		\label{fig_2}
	\end{figure}

	Let $M$ point sets donate by $X=\left\{ {{x}_{i}} \right\}_{i=1}^{M}$ and ${{x}_{i}}=\left[ {{x}_{i,1}},{{x}_{i,2}},\ldots ,{{x}_{i,l}},\ldots ,{{x}_{i,{{N}_{i}}}} \right]$, where ${{N}_{i}}$ is the number of data points in the $i$th point set. Given one data point ${{x}_{i,l}}$ of the $i$th point set, its nearest neighbor ${{x}_{j,c\left( j,l \right)}}$ can be searched for each other point set. Considering noise and outlier, it is reasonable to assume that the data point ${{x}_{i,l}}$ is generated from the t-distribution with the centroid ${{x}_{j,c\left( j,l \right)}}$.
	To utilize all available information contained in each other data set, we further suppose that the data point ${{x}_{i,l}}$ is generated from the StMM, where $\left\{ {{x}_{j,c\left( j,l \right)}} \right\}_{j=1,j\ne i}^{M}$ are regarded as its component centroids. As shown in Fig.~\ref{fig_2}, the data point ${{x}_{i,l}}$ is drawn from the unique mixture model consisting of $\left( M-1 \right)$ t-distribution components, where its nearest neighbors in other point sets are regarded as the t-distribution centroids.

	Subsequently, we define a kinematic model as $X=X\left( \Phi  \right)$ and $\Phi =\left\{ {\mathbf{R}_{i}},{{t}_{i}} \right\}_{i=1}^{M}$, which contains rotation matrix $\mathbf{R}$ and translation vector $t$ in the rigid transformation. To simplify the calculation, each t-distribution component is assigned with the equal membership probability, degrees of freedom $v$ and covariance. What's more, the covariance is set to be isotropic, i.e. $\Sigma ={{\sigma }^{2}}{\mathbf{I}_{3}}$, where ${\mathbf{I}_{3}}$ denotes the 3D identity matrix. Under these assumptions, it is easy to define the formula for the joint probability of data point ${{x}_{i,l}}$ as:
	\begin{equation}
	    P\left( {{x}_{i,l}} \right)=\sum\limits_{j\ne i}^{M}{\frac{1}{M-1}{{f}_{T}}\left( {{x}_{i,l}}\left( {{\Phi }_{i}} \right);{{x}_{j,c\left( j,l \right)}},{{\sigma }^{2}},v \right)},
	    \label{eq_5}
	\end{equation}
	Different from GMM based models, there is no need to add a uniform distribution to account for outliers. This is mainly because the t-distribution tolerates to noises and outliers.

	\section{Expectation-maximization algorithm}
	The parameter set $\Theta =\left\{ \Phi ,{\Sigma} \right\}$ can be optimized by maximizing  Eq.~(\ref{eq_5}). However, there is no closed-form solution to directly optimize this problem due to incomplete data. Therefore, we will define maximum likelihood function and optimize it by the EM algorithm to estimate the parameter set.
	
	\subsection{Maximum likelihood estimation of StMM}
	To apply the EM algorithm, it requires to define the complete data set as follows:
	\begin{equation}
	    {{X}_{compelete}}={{\left( X,Z,U \right)}^{T}}
	    \label{eq_6}
	\end{equation}
	where $X=\left\{ {{x}_{i,l}}\left| l\in \left[ 1,\ldots ,{{N}_{i}} \right] \right. \right\}_{i=1}^{M}$ denote multi-view point sets being registered, $Z=\left\{ {{z}_{i,l}}\left| l\in \left[ 1,\ldots ,{{N}_{i}} \right] \right. \right\}_{i=1}^{M}$ and $U=\left\{ {{u}_{i,l}}\left| l\in \left[ 1,\ldots ,{{N}_{i}} \right] \right. \right\}_{i=1}^{M}$ indicate two hidden variable sets. More specifically, the hidden variable ${{z}_{i,l}}=c\left( j,l \right)$ means that the data point ${{x}_{i,l}}$ is drawn from the t-distribution with the centroid ${{x}_{j,c\left( j,l \right)}}$. In addition, the hidden variable ${{u}_{i,j}}$ is utilized to adjust the size of the covariance in the t-distribution, which generates the data point ${{x}_{i,l}}$. According to \cite{peel2000robust}, the following relationships can be derived from (Eq.~(\ref{eq_2})):
	\begin{equation}
	    {{u}_{i,l}}\left| {{z}_{i,l}}=c\left( j,l \right) \right.\sim \mathcal{G}\left( {v}/{2}\;,{v}/{2}\; \right)
	    \label{eq_7}
	\end{equation}
	\begin{equation}
	    {{x}_{i,l}}\left| {{u}_{i,l}},{{z}_{i,l}}=c\left( j,l \right) \right.\sim \mathcal{N}\left( {{x}_{j,c\left( j,l \right)}},{{{\sigma }^{2}}}/{u}\; \right)
	    \label{eq_8}
	\end{equation}
	
Based on the Bayesian formula, we can formulate the joint probability density function as:
	\begin{align}
	    P\left( X,U,Z;\Theta  \right) &=P\left( X\left| U,Z;\Theta  \right. \right)P\left( U,Z;\Theta  \right) \notag\\ 
	    &=P\left( X\left| U,Z;\Theta  \right. \right)P\left( U\left| Z;\Theta  \right. \right)P\left( Z;\Theta  \right)
	    \label{eq_9}
	\end{align}
As equal membership probability has been assigned to each t-distribution component, $P\left( Z;\Theta  \right)$ denotes a constant term. Therefore, the log-likelihood function $L\left( \Theta  \right)$ of complete data is denoted as follows:
	\begin{align}
	   & L\left( \Theta  \right) \notag\\ 
	   & =\log P\left( X,U,Z;\Theta  \right) \notag\\ 
	   & =\log P\left( X\left| U,Z;\Theta  \right. \right)+\log P\left( U\left| Z;\Theta  \right. \right) \notag\\ 
	   & =\sum\limits_{i,l,j}{\left[ -\frac{d}{2}\log 2\pi -\frac{d}{2}\log {{\sigma }^{2}}+\frac{d}{2}\log {{u}_{i,l,j}} \right.}\left. -\frac{1}{2}{{u}_{i,l,j}}{{\Delta }_{i,l,j}}^{2} \right] \notag\\ 
	   & +\sum\limits_{i,l,j}{\left[ \frac{v}{2}\log \left( \frac{v}{2} \right)-\log \left( \Gamma \left( \frac{v}{2} \right) \right)+ \right.\frac{v}{2}\left( \log {{u}_{i,l,j}}-{{u}_{i,l,j}} \right)} \notag\\ 
	   & \left. -\log \left( {{u}_{i,l,j}} \right) \right]
	   \label{eq_10}
	\end{align}
where ${{\Delta }_{i,l,j}}^{2}=\frac{\left\| {{x}_{i,l}}\left( {{\Phi }_{i}} \right)-{{x}_{j,c\left( j,l \right)}} \right\|_{2}^{2}}{{{\sigma }^{2}}}$ denotes the squared Mahalanobis distance and the symbol ${{x}_{i,l}}\left( {{\Phi }_{i}} \right)= \mathbf{R}_{i}{{x}_{i,l}}+ t_i$ indicates imposing the rigid transformation $\{\mathbf{R}_{i},t_i\}$ on the data point ${{x}_{i,l}}$. Subsequently, the likelihood function $L\left( \Theta  \right) $ should be maximized by the EM algorithm, which alternatively implements the E-step and M-step to optimize all rigid transformations for the multi-view registration.
	
\subsection{E-step}
Given all point sets $X$ and the current estimated parameters $\Theta^{(k-1)}$, the E-step calculates the expected value of $L(\Theta)$. To define the StMM, the t-distribution centroids $\left\{ {{x}_{j,c\left( j,l \right)}} \right\}_{j=1,j\ne i}^{M}$ should be updated by the establishment of point correspondences. As all point sets are constantly moving during registration, the point correspondences are established between the $i$th point set to other aligned point sets:
\begin{equation}
	    c^{(k)}\left( j,l \right)=\underset{h\in \left[ 1,2,\ldots ,{{N}_{j}} \right]}{\mathop{\min }}\,{{\left\| {{x}_{i,l}}\left( {{\Phi }_{i}^{(k-1)}} \right)-{{x}_{j,h}} \right\|}_{2}}.
	    \label{eq_11}
	\end{equation}
Eq.~(\ref{eq_11}) denotes the NN search problem, which has been efficiently solved by the $k$-d tree based method \cite{nuchter2007cached}. Once all centroids are determined for the StMM, a modified set of posterior probabilities $P_{i,l,j}^{*}$ can be calculated by multiplying the conditional expectations of $Z=\left\{ {{z}_{i,l}}\left| l\in \left[ 1,\ldots ,{{N}_{i}} \right] \right. \right\}_{i=1}^{M}$ and $U=\left\{ {{u}_{i,l}}\left| l\in \left[ 1,\ldots ,{{N}_{i}} \right] \right. \right\}_{i=1}^{M}$, where $P_{i,l,j}^{*}$ gives the robust correspondence probabilities between the data point ${{x}_{i,l}}$ and each t-distribution centroid ${{x}_{j,c\left( j,l \right)}}$. The conditional expectation of ${{z}_{i,l}}$ is calculated as follows:
	\begin{align}
	    &{{E}_{{{\Theta }^{(k-1)}}}}\left( {{z}_{i,l}}\left| {{x}_{i,l}} \right. \right) \notag\\
	    & =P_{i,l,j}^{\left( k \right)}=\frac{{{f}_{T}}\left( {{x}_{i,l}}\left( {{\Phi }_{i}} \right);{{x}_{j,c\left( j,l \right)}},{{\Sigma }^{(k-1)}},v \right)}{\sum\nolimits_{h=1,h\ne i}^{M}{{{f}_{T}}\left( {{x}_{i,l}}\left( {{\Phi }_{i}} \right);{{x}_{h,c\left( h,l \right)}},{{\Sigma }^{(k-1)}},v \right)}}.
	    \label{eq_12}
	\end{align}
	
According to \cite{liu1995ml}, the degree of freedom for $\mathcal{X}_{d}^{2}$ distribution is $u{{\Delta }_{i,l,j}}$. Therefore,
the log-likelihood of ${{u}_{i,l}}$ is:
	\begin{equation}
	    L\left( {{u}_{i,l}}\left| {{x}_{i,l}},{{z}_{i,l}}=c^{(k)}\left( j,l \right) \right. \right)\propto Gamma\left( \frac{d}{2},\frac{{{\Delta }_{i,l,j}}}{2} \right).
	    \label{eq_13}
	\end{equation}
As the Gamma distribution denotes the conjugate prior distribution for ${{u}_{i,l}}$, the conditional distribution of ${{u}_{i,l}}$ presented in by Eq.~(\ref{eq_7}) and (\ref{eq_13}) is:
	\begin{equation}
	    {{u}_{i,l}}\left| {{x}_{i,l}},{{z}_{i,l}}=c^{(k)}\left( j,l \right) \right.\sim Gamma\left( \frac{v+d}{2},\frac{v+{{\Delta }_{i,l,j}}}{2} \right).
	    \label{eq_14}
	\end{equation}
From Eq.~(\ref{eq_14}), the conditional expectation of ${{u}_{i,l}}$ can be determined as:
	\begin{equation}
	    {{E}_{{{\Theta }^{(k-1)}}}}\left( {{u}_{i,l,j}}\left| {{x}_{i,l}},{{z}_{i,l}}=c^{(k)}\left( j,l \right) \right. \right)=U_{i,l,j}^{\left( k \right)}=\frac{{{v}_{j}}+d}{{{v}_{j}}+{{\Delta }_{i,l,j}}^{2}}.
	    \label{eq_15}
	\end{equation}
	
From Eq.~(\ref{eq_12}) and Eq.~(\ref{eq_15}), it is reasonable to formulate the robust posterior probabilities $P_{i,l,j}^{*}$ as:
	\begin{align}
	    &{{E}_{{{\Theta }^{(k-1)}}}}\left( {{z}_{i,l}}\left| {{x}_{i,l}} \right. \right){{E}_{{{\Theta }^{(k-1)}}}}\left( {{u}_{i,l,j}}\left| {{x}_{i,l}},{{z}_{i,l}}=c^{(k)}\left( j,l \right) \right. \right)\notag\\
	    &=P_{i,l,j}^{*\left( k \right)}=P_{i,l,j}^{\left( k \right)}U_{i,l,j}^{\left( k \right)}.
	    \label{eq_16}
	\end{align}
Once the posterior probabilities $P_{i,l,j}^{*}$ is computed from the current estimated parameters, the model parameters $\Theta$ should be further optimized by the M-step. 
	
\subsection{M-step}
	The M-step requires to maximize the log-likelihood expectation over the parameter set $\Theta =\left\{ \Phi ,\Sigma  \right\}$. Under the posterior probabilities $P_{i,l,j}^{*}$,
	the log-likelihood expectation can be reformulated as:
   \begin{align}
	& Q\left( {{\Theta }^{\left( k \right)}}\left| {{\Theta }^{\left( k-1 \right)}} \right. \right)=E\left( L\left( \Theta  \right) \right) \notag\\ 
	& =\sum\limits_{Z}{\sum\limits_{i,l,j}{P_{i,l,j}^{\left( k \right)}\left[ \frac{v}{2}\log \left( \frac{v}{2} \right)-\log \left( \Gamma \left( \frac{v}{2} \right) \right) \right.}} \notag\\
	& \left. +\frac{v}{2}\left( \log U_{i,l,j}^{\left( k \right)}-U_{i.l.j}^{\left( k \right)} \right)-\log \left( U_{i,l,j}^{\left( k \right)} \right) \right] \notag\\ 
	& +\sum\limits_{Z}{\sum\limits_{i,l,j}{P_{i,l,j}^{\left( k \right)}\left[ -\frac{d}{2}\log 2\pi -\frac{d}{2}\log {{\sigma }^{2}}+\frac{d}{2}\log U_{i,l,j}^{\left( k \right)} \right.}} \notag\\ 
	& \left. -\frac{1}{2}U_{i,l,j}^{\left( k \right)}{{\Delta }_{i,l,j}}^{2} \right].  
	\label{eq_17}
	\end{align}
  Obviously, there are many model parameters included in Eq. (\ref{eq_17}), which is difficult to be directly optimized.
  Fortunately, these model parameters can be alternatively optimized. More specifically, the rigid transformation $\Phi^{(k)}$ can be updated by fixing the t-distribution covariacne $\sigma^2$: 
	\begin{equation}
	    \left\{ \begin{matrix}
	    \underset{{\mathbf{R}_{i}},{{t}_{i}}}{\mathop{\arg \min }}\,\left( \sum\limits_{Z}{\sum\limits_{i,l,j}{P_{i,l,j}^{*\left( k \right)}\left\| {{x}_{i,l}}\left( {{\Phi }_{i}^{(k)}} \right)-{{x}_{j,c^{(k)}\left( j,l \right)}} \right\|_{2}^{2}}} \right)  \\
	    s.t.\begin{array}{*{35}{l}}
	     \\
	    \end{array}\mathbf{R}_{i}^{T}{\mathbf{R}_{i}}=\mathbf{I}^3\begin{array}{*{35}{l}}
	    {}  \\
	    \end{array}and\begin{array}{*{35}{l}}
	    {}  \\
	    \end{array}\left| {\text{R}_{i}} \right|=1,\begin{array}{*{35}{l}}
	    {}  \\
	    \end{array}\forall i\in \left[ 1,\ldots ,M \right].  \\
	    \end{matrix} \right.
	    \label{eq_18}
	\end{equation}
	Eq. (\ref{eq_18}) consists several weighted least squares estimation sub-problems. Given $\left\{ P_{i,l,j}^{\left( k \right)},U_{i,l,j}^{\left( k \right)},v,{{\Sigma }^{(k-1)}} \right\}$, they can be sequentially optimized by the Singular value decomposition (SVD) based method \cite{nuchter2010study} to update each rigid transformation. 
	
	After the update of ${{\Phi }}$, the covarince matrix $\Sigma$ can be updated by taking the partial derivatives of $Q$ with respect to ${{\sigma }^{2}}$ and setting it to 0. That is
	\begin{equation}
	    {\Sigma}^{(k)}={\sigma}^{2}\mathbf{I}_3,
	\end{equation}
	where
	\begin{equation}
	    {\sigma}^{2}=\frac{\sum\nolimits_{i,l,j}{P_{i,l,j}^{*\left( k \right)}\left\| {{x}_{i,l}}\left( {{\Phi }_{i}^{(k)}} \right)-{{x}_{j,c^{(k)}\left( j,l \right)}} \right\|_{2}^{2}}}{d\sum\nolimits_{i,l,j}{P_{i,l,j}^{\left( k \right)}}}.
	    \label{eq_19}
	\end{equation}
	
Obviously, the number of StMM components is automatically determined by
the number of point sets being registered. Therefore, there is no trade-off between registration accuracy and efficiency in the proposed method. Besides, all t-distribution components are defined by the NN search method, so the proposed method is efficient to optimize all other model parameters. Compared with GMM based methods, StMM is robust to heavy-tail noises as well as outliers, which make the propose method be able to obtain desired registration results.  

	\begin{algorithm}[!t]
	    \caption{}
	    \label{Algorithm_1}
	    \renewcommand{\algorithmicrequire}{\textbf{Input:}}
        \renewcommand{\algorithmicensure}{\textbf{Output:}}
	    \begin{algorithmic}[1]
	        \REQUIRE Point sets $X=\left\{ {{x}_{i}} \right\}_{i=1}^{M}$, maximum iteration $K=300$,\\ initial guesses ${{\Theta }^{0}}$, $\varepsilon =0.0005$.
            \ENSURE $\Phi =\left\{ {\mathbf{R}_{i}},{{t}_{i}} \right\}_{i=1}^{M}$
	        \STATE $k=0 $;
	        \REPEAT   
            \STATE $k=k+1$;
            \FOR {$\left( i=1:M \right)$}  
            \STATE E-step: 
            \STATE Build correspondence $\left\{ x_{i,l},x_{j,c^{(k)}{{\left( j,l \right)}}} \right\}$ by Eq. (11);
            \STATE Compute $P_{i,l,j}^{*\left( k \right)}$ by Eq. (16);
            \STATE M-step:
            \STATE Update ${{\Phi }^{\left( k \right)}}$ by solve Eq. (18);
            \STATE Update ${\Sigma}^{(k)}$ by Eq. (19);
            \ENDFOR
            \UNTIL{$\left( \frac{1}{M}\left| Q\left( {{\Theta }^{\left( k \right)}} \right)-Q\left( {{\Theta }^{\left( k-1 \right)}} \right) \right|<\varepsilon  \right)$ or $\left( k>K \right)$}
	    \end{algorithmic}
	\end{algorithm}

			\begin{table*}[!t]
	    \caption{Information statistics of object data sets in experiment}
	    \label{Tabel1}
	    \centering
	    \newcolumntype{C}[1]{>{\centering\arraybackslash}p{#1}}
	    \begin{tabular}{C{2cm}|C{2cm}|C{2cm}|C{2cm}|C{2cm}|C{2cm}|C{2cm}}
	         \toprule
	         & {Angle} & {Armadillo} & {Bunny} & {Buddha} & {Dragon} & {Hand}\\
	         \midrule
	         Scans & 36 & 12 & 10 & 15 & 15 & 36\\
	         \midrule
	         Point & 2347854 & 307625 & 362272 & 1099005 & 469193 & 1605575\\
	         \bottomrule
	    \end{tabular}
	\end{table*}

	\subsection{Implementation}
	The proposed method utilizes the EM algorithm to achieve multi-view registration. Similar to other EM based method, our method requires initial model parameters ${\Theta}^{0} =\left\{ {\Phi}^{0} ,{\Sigma }^{0} \right\}$ to start the optimization. Usually, 
	${{\Phi }^{0}}$ is provided by other coarse registration methods. Empirically, we use the average point resolution $d_r$ of all point sets to initialize ${\Sigma }^{0}$, such as ${\Sigma }^{0}= (d_r)^2\rm{I}_3$. Based on this initialization, the EM algorithm optimizes all model parameters and terminates optimization once the iteration number reaches the maximum step $K=300$ or the objective function change is less than threshold $\varepsilon=0.0005$. According to above-mentioned descriptions, the proposed method can be summarized in Algorithm 1. 	
	
	\section{Experiments}
	To illustrate its performance, the proposed method was tested on six data sets, where four data sets were taken from Stanford 3D Scanning Repository \cite{levoy2005stanford}, and the other two were provided by Torsello \cite{torsello2011multiview}. In addition to scan data, these data sets also include the ground truth of rigid transformations $\left\{ {\mathbf{R}_{g,i}},{{t}_{g,i}} \right\}_{i=1}^{M}$, which can be used to evaluate the accuracy of multi-view registration results. Some details of these data sets are displayed in Table.~\ref{Tabel1}, which includes the number of scans and the total number of data points. For the comparison, the proposed method is compared with three state-of-the-art methods abbreviated as JRMPC \cite{evangelidis2017joint}, TMM \cite{ravikumar2018group} and EMPMR \cite{zhu2020registration}. More	specifically, 1) JRMPC: It assumes that all data points are realizations of one central GMM. 2) TMM: It assumes that all data points are generated by one central StMM. 
	3) EMPMR: It supposes that each point is drawn from one unique GMM, where its NNs in other point sets are regarded as Gaussian centroids with equal covariance and membership probabilities. For the efficiency, all data sets were down-sample around to 2000 points per scan for multi-view registration. Considering the fairness, all compared method utilizes the same parameter setting for each data set. For the evaluation of accuracy, rotation error and translation error are defined as:
	\begin{equation}
	{{e}_\mathbf{R}}=\frac{1}{M}\sum\nolimits_{i=1}^{M}{\arccos \left( \frac{\text{tr}\left( {\mathbf{R}_{m,i}}{{\left( {\mathbf{R}_{g,i}} \right)}^{T}} \right)-1}{2} \right)}
	\end{equation}
	and 
	\begin{equation}
		{{e}_{t}}=\frac{1}{M}{{\sum\nolimits_{i=1}^{M}{\left\| {{t}_{m,i}}-{{t}_{g,i}} \right\|}}_{2}},
\end{equation}
	where $\left\{ {\mathbf{R}_{m,i}},{{t}_{m,i}} \right\}_{i=1}^{M}$ denotes the rigid transformations estimated by some registration methods. All compared methods were implemented on Matlab and these codes were run on a four-core 3.0 GHz computer with 16 GB of memory.
	
	\subsection{Parameter sensitivity and Convergence}
\begin{figure}[!t]
		\centering
		\subfigure[]{
		\centering
		\includegraphics[width=0.85\columnwidth]{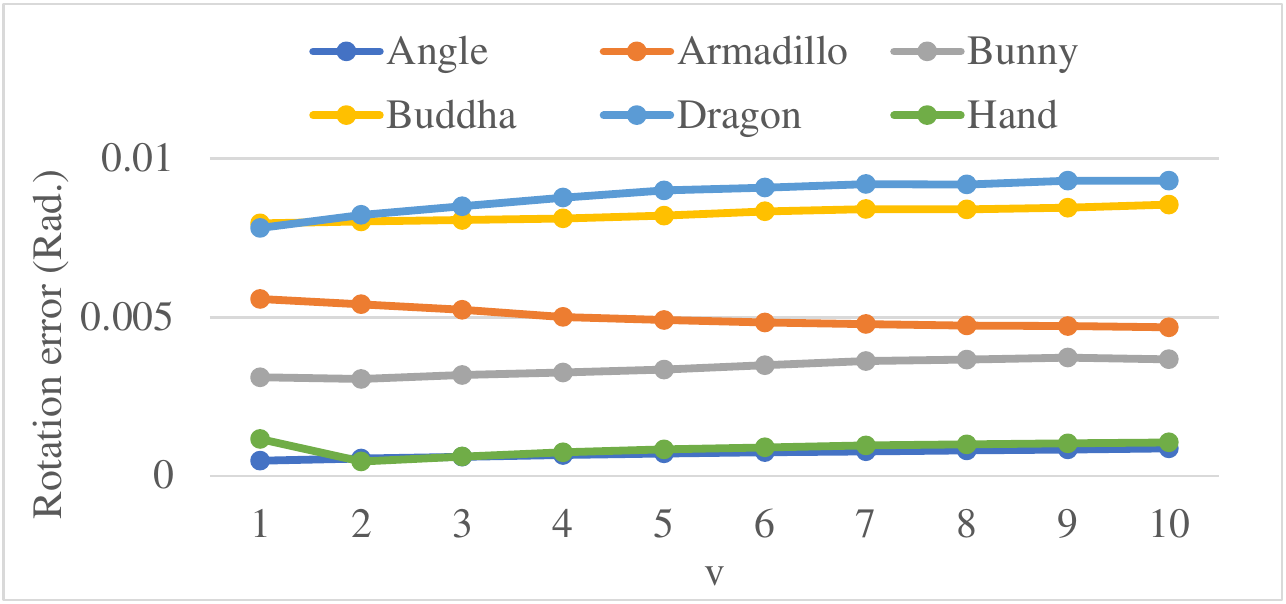}
		}
		\subfigure[]{
		\centering
		\includegraphics[width=0.85\columnwidth]{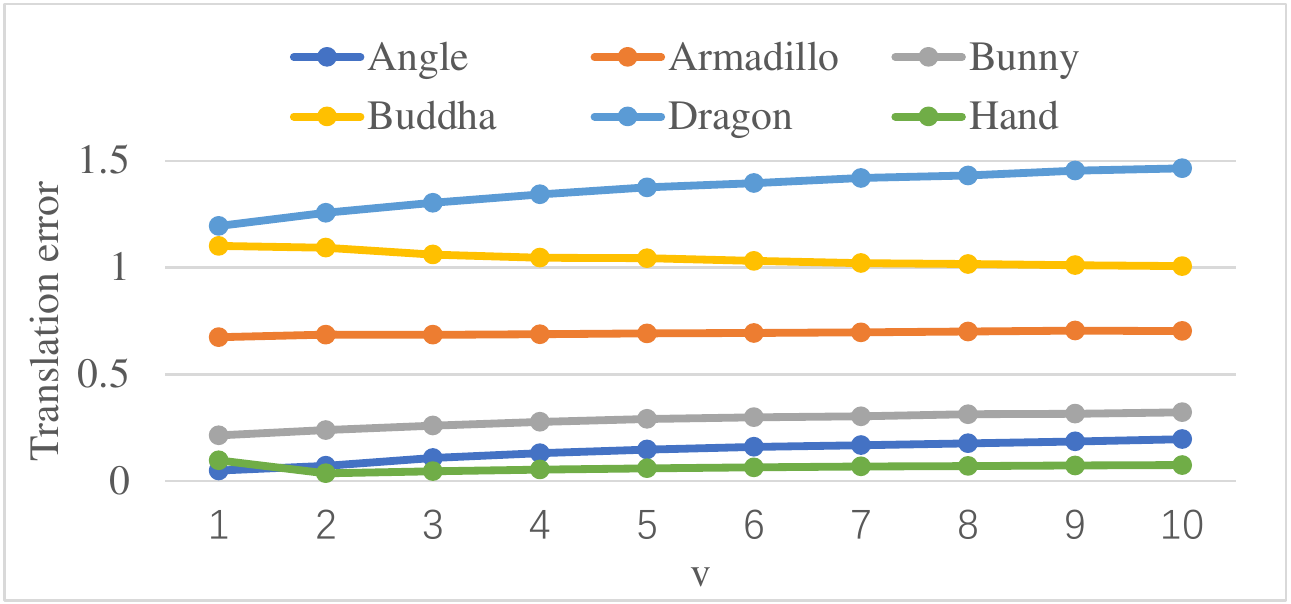}
		}
		\caption{The influence of $v$ to the registration performance of our method tested six data sets. (a) Rotation errors. (b) Translation errors.}
		\label{fig_3}
	\end{figure}

   In the proposed method, there is only one hyper-parameter $v$, which controls the tail of t-distribution. Accordingly, the proposed method is tested on six data sets under different values of $v$, so as to view its effect on registration performance. Experimental results are reported in the form of registration errors. Fig.~\ref{fig_3} illustrates registration errors of different data sets under varied values of $v$. As shown in  Fig.~\ref{fig_3}, once the hyper-parameter $v$ is chosen in a suitable range, i.e., from 2 to 10, it has small influence on the registration performance of the proposed method. What's more, the proposed method is relatively insensitive to this hyper-parameter. This allow us to easily apply the proposed method with less effort for parameter tuning. As shown in Fig.~\ref{fig_2}, the smaller $v$ is, the heavier the tail of t-distribution is. However, too small $v$ will make the proposed method be difficult to distinguish data and noises. Accordingly, 
   we set $v=3$ in following experiments.
 
\begin{figure}
		\centering
		\includegraphics[width=0.85\columnwidth]{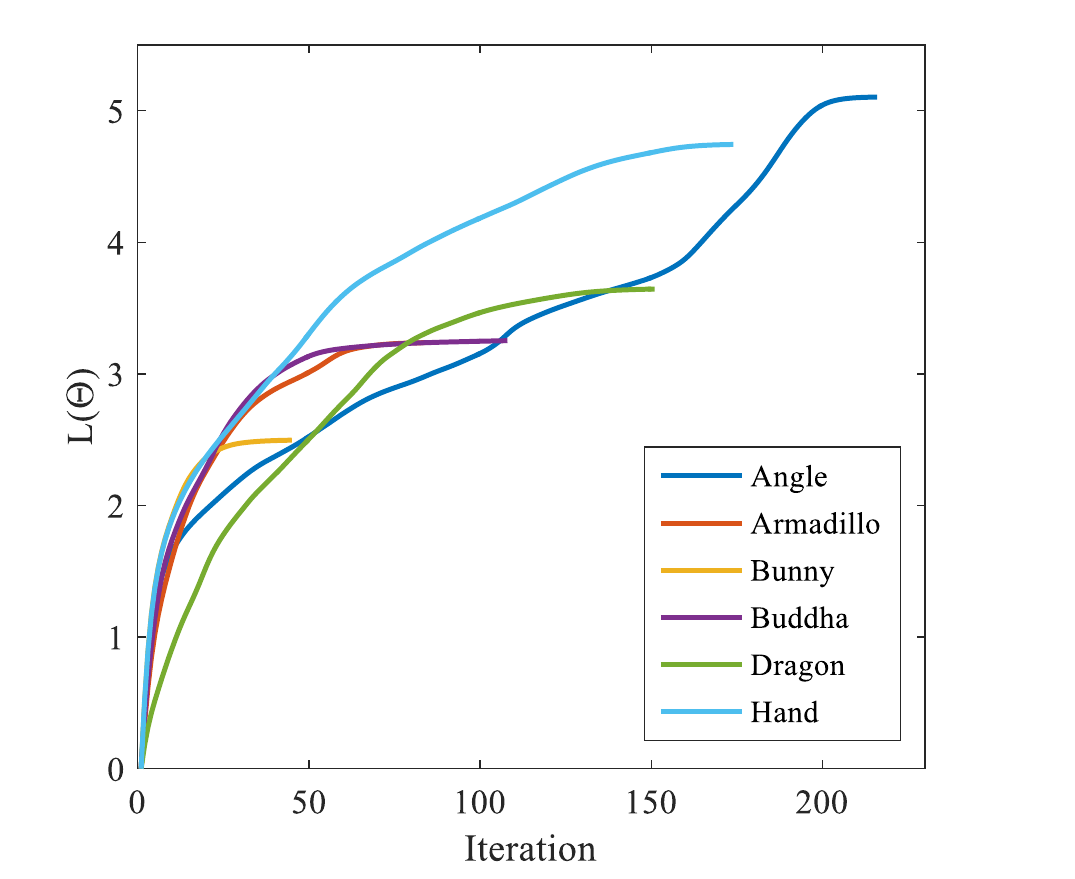}		\caption{Convergence illustration of our method tested on six data sets. For the display convenience, we illustrate the log-likelihood value subtracted by initial value for each data set.}
		\label{fig_4}
	\end{figure}

		\begin{table*}[!t]
	    \centering
	    \caption{The registration errors of all methods of different levels of initial rotation on Stanford Bunny (mean$\pm$std.),\\ where the numbers in bold means the best performance on each data set.}
	    \begin{tabular}{c|c|c|c|c|c|c}
	    \toprule
	    Method& ~ & $\left[ -0.01,0.01 \right]$ rad. & $\left[ -0.02,0.02 \right]$ rad. & $\left[ -0.03,0.03 \right]$ rad. & $\left[ -0.04,0.04 \right]$ rad. & $\left[ -0.05,0.05 \right]$ rad. \\
	    \midrule
		\multirow{2}{*}{JRMPC} & ${{e}_{R}}$ & $0.0195\pm 0.0042$ & $0.0519\pm 0.0081$ & $0.0790\pm 0.0102$ & $0.1084\pm 0.0119$ & $0.1382\pm 0.0167$ \\
		& ${{e}_{t}}$ & $1.3937\pm 0.2799$ & $2.5948\pm 0.4775$ & $2.3208\pm 0.5213$ & $2.0454\pm 0.3736$ & $1.6979\pm 0.2920$ \\
		\midrule
		\multirow{2}{*}{TMM} & ${{e}_{R}}$ & $0.0133\pm 0.0023$ & $0.0185\pm 0.0047$ & $0.0216\pm 0.0043$ & $0.0247\pm 0.0076$ & $0.0483\pm 0.0195$ \\
		& ${{e}_{t}}$ & $1.1289\pm 0.2088$ & $1.6244\pm 0.2936$ & $1.9049\pm 0.3441$ & $2.1084\pm 0.5682$ & $3.8180\pm 1.3794$ \\
	    \midrule
	    \multirow{2}{*}{EMPMR} & ${{e}_{R}}$ & $0.0040\pm 0.0004$ & $0.0070\pm 0.0057$ & $0.0650\pm 0.0319$ & $0.1061\pm 0.0287$ & $0.1378\pm 0.0175$ \\
	    & ${{e}_{t}}$ & $0.3899\pm 0.0255$ & $0.6372\pm 0.5080$ & $3.6076\pm 1.6647$ & $4.8930\pm 1.5960$ & $5.4171\pm 1.5450$ \\
	    \midrule
	    \multirow{2}{*}{Our Method} & ${{e}_{R}}$ & \bm{$0.0036\pm 0.0005$} & \bm{$0.0039\pm 0.0008$} & \bm{$0.0039\pm 0.0004$} & \bm{$0.0059\pm 0.0094$} & \bm{$0.0171\pm 0.0330$} \\
	    & ${{e}_{t}}$ & \bm{$0.3470\pm 0.0369$} & \bm{$0.3557\pm 0.0596$} & \bm{$0.3700\pm 0.0305$} & \bm{$0.5379\pm 0.7980$} & \bm{$1.0927\pm 1.7275$} \\
	    \toprule
	    \end{tabular}
	    \label{table_2}
	\end{table*}
	
	\begin{table*}[!t]
	    \centering
	    \caption{The registration errors of all methods of different levels of initial translation on Stanford Bunny (mean$\pm$std.),\\ where the numbers in bold means the best performance on each data set.}
	    \begin{tabular}{c|c|c|c|c|c|c}
	    \toprule
	    Method& ~ & $\left[ -2.4,2.4 \right]\times {{d}_{r}}$ & $\left[ -3.2,3.2 \right]\times {{d}_{r}}$ & $\left[ -4,4 \right]\times {{d}_{r}}$ & $\left[ -4.8,4.8 \right]\times {{d}_{r}}$ & $\left[ -5.6,5.6 \right]\times {{d}_{r}}$ \\
	    \midrule
	    \multirow{2}{*}{JRMPC} & ${{e}_{R}}$ & $0.0208\pm 0.0041$ & $0.0224\pm 0.0062$ & $0.0212\pm 0.0045$ & $0.0185\pm 0.0029$ & $0.0172\pm 0.0031$ \\
		& ${{e}_{t}}$ & $2.9940\pm 0.4588$ & $4.8254\pm 0.8405$ & $6.5497\pm 1.0927$ & $8.2998\pm 1.0634$ & $9.7766\pm 1.1661$ \\
		\midrule
		\multirow{2}{*}{TMM} & ${{e}_{R}}$ & $0.0151\pm 0.0029$ & $0.0171\pm 0.0028$ & $0.0204\pm 0.0048$ & $0.0185\pm 0.0043$ & $0.0221\pm 0.0032$ \\
		& ${{e}_{t}}$ & $1.3953\pm 0.3070$ & $1.6038\pm 0.3340$ & $1.8576\pm 0.4039$ & $1.7792\pm 0.4080$ & $2.2833\pm 0.6973$ \\
		\midrule
		\multirow{2}{*}{EMPMR} & ${{e}_{R}}$ & $0.0051\pm 0.0045$ & $0.0096\pm 0.0084$ & $0.0192\pm 0.0110$ & $0.0304\pm 0.0131$ & $0.0366\pm 0.0102$ \\
		& ${{e}_{t}}$ & $0.5803\pm 0.6770$ & $1.2345\pm 1.3153$ & $2.6259\pm 1.6753$ & $5.2263\pm 2.0695$ & $7.2838\pm 3.0659$ \\
		\midrule
		\multirow{2}{*}{Our Method} & ${{e}_{R}}$ & \bm{$0.0037\pm 0.0005$} & \bm{$0.0039\pm 0.0004$} & \bm{$0.0038\pm 0.0005$} & \bm{$0.0040\pm 0.0006$} & \bm{$0.0069\pm 0.0123$} \\
		& ${{e}_{t}}$ & \bm{$0.3621\pm 0.0357$} & \bm{$0.3745\pm 0.0393$} & \bm{$0.3752\pm 0.0465$} & \bm{$0.3814\pm 0.0420$} & \bm{$0.8381\pm 1.4268$} \\
		\toprule
	    \end{tabular}
	    \label{table_3}
	\end{table*}
	
	To view its convergence properties, Fig.~\ref{fig_4} illustrates the log-likelihood value of the proposed method at each iteration. As shown in Fig.~\ref{fig_4}, the proposed method can always converge quickly due to the closed-form optimization solution. But the required iteration number is increased with the number of scans involved in the multi-view registration. 
	This is because more rigid transformations should be optimized for the large data sets, which is inevitable to reduce the convergence speed of the proposed method.
	However, this limitation is also shared by most of other multi-view registration methods.

	\subsection{Effect of initial rigid transformations}
	To illustrate its robustness to initialization, the proposed method was tested on Stanford Bunny with different initial rigid transformations and compared with other three methods. Since the rigid transformation includes the rotation matrix and translation vector, rotation angles or translation variables is drawn from different uniform distributions with varying intervals, respectively. Further, the other one is assigned with random typical values, so as to generate the disturbance quantity of rigid transformation $(\Delta \mathbf{R}_i,\Delta t_i)$. Subsequently, it is easy to obtain the initial rigid transformation by adding the disturbance quantity to the ground truth, such as $\mathbf{R}_{i}^{0}=\Delta {\mathbf{R}_{i}}\cdot {\mathbf{R}_{i}}$ and $t_{i}^{0}=\Delta {{t}_{i}}+{{t}_{i}}$. To eliminate randomness, each method was tested on one level of rigid transformation disturbance by 20 times. Experimental results are reported in the form of registration errors, which are displayed in Table.~\ref{table_2} and Table.~\ref{table_3}, where $d_r$ denotes the average point resolution for all scans. As shown in Table.~\ref{table_2} and Table.~\ref{table_3}, the proposed method is the most robust one to initial rigid transformations. More specifically, when the level of disturbance quantity is low, all compared methods can obtain promising results. With the increase of disturbance quantity, our method can still achieve multi-view registration with good accuracy, but other three methods fluctuate drastically and are unable to obtain promising results.

Given the large disturbance quantity, JRMPC and TMM are more likely trapped into local minimum due to massive model parameters, which are required to be optimized. Compared with JRMPC and TMM, both EMPMR and our method require optimizing less model parameters. Therefore, both of them are more likely to obtain promising results under good initialization. However, EMPMR suffers from the weakness of GMM in the noise with heavy-tail. Besides, it utilizes one uniform distribution to model outliers and requires to manually tune the weight, which controls the effect of the uniform distribution component with respective to all other Gaussian components. With the increase of disturbance quantity, its registration performance will drastically reduce. While, our method utilizes the StMM to formulate the multi-view registration problem, it takes full account of the noise with heavy-tail as well as outliers, so there is no need to use uniform distribution for extra modeling outliers. Therefore, our method can obtain promising registration results even under large disturbance quantity. Overall, the proposed method is very robust to initialization.

    \begin{table*}[!t]
	    \centering
	    \caption{The registration errors of each method having the same initial rigid transformation,\\ where the numbers in bold means the best performance on each data set.}
	    \begin{tabular}{c|c|c|c|c|c|c|c}
	    \toprule
	    Method & ~ & Angel (dm) & Armadillo (mm) & Bunny (mm) & Buddha (mm) & Dragon (mm) & Hand (dm) \\
	    \midrule
	    \multirow{2}{*}{Initial} & ${{e}_{R}}$ & 0.0275 & 0.0243 & 0.0239 & 0.0273 & 0.0552 & 0.0582 \\
		& ${{e}_{t}}$ & 2.0690 & 3.7207 & 2.1260 & 2.8440 & 5.3504 & 0.4986 \\
		\midrule
		\multirow{2}{*}{JRMPC} & ${{e}_{R}}$ & 0.0308 & 0.0170 & 0.0176 & 0.0384 & 0.0555 & 0.0079 \\
		& ${{e}_{t}}$ & 6.8760 & 1.3235 & 1.5463 & 1.8648 & 5.0833 & 0.8233 \\
		\midrule
		\multirow{2}{*}{TMM} & ${{e}_{R}}$ & 0.0066 & 0.0263 & 0.0119 & 0.0201 & 0.0182 & 0.0055 \\
		& ${{e}_{t}}$ & 2.9481 & 2.5593 & 0.7506 & 1.1965 & \bf{1.5088} & 0.8376 \\
		\midrule
		\multirow{2}{*}{EMPMR} & ${{e}_{R}}$ & 0.0008 & 0.0206 & 0.0035 & 0.0083 & 0.0624 & 0.0012 \\
		& ${{e}_{t}}$ & 0.1747 & 1.7521 & 0.3439 & 1.3030 & 4.4142 & 0.0836 \\
		\midrule
		\multirow{2}{*}{Our Method} & ${{e}_{R}}$ & \bf{0.0006} & \bf{0.0054} & \bf{0.0032} & \bf{0.0065} & \bf{0.0130} & \bf{0.0006} \\
		& ${{e}_{t}}$ & \bf{0.1079} & \bf{0.7342} & \bf{0.2599} & \bf{1.1553} & 1.5224 & \bf{0.0458} \\
	    \toprule
	    \end{tabular}
	    \label{table_4}
	\end{table*}
	
	\begin{figure*}
		\centering
		\includegraphics[width=1.9\columnwidth]{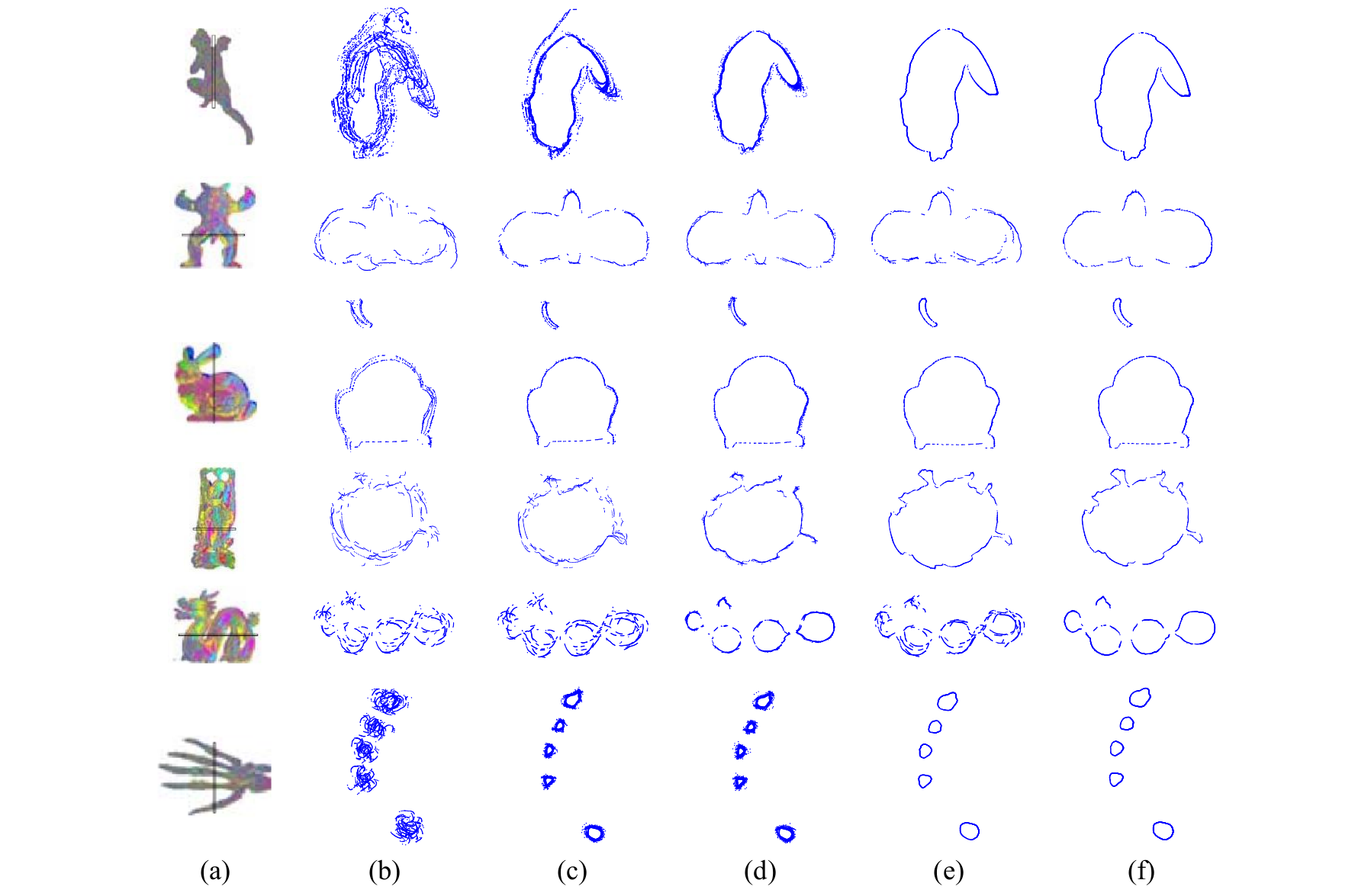}
		\caption{Registration results displayed in the form of cross-section, where the corresponding regions are indicated on the aligned 3D models. (a) Aligned 3D models. (b) Initial results. (c) JRMPC results. (d) TMM results. (e) EMPRM Results . (f) Our results.}
		\label{fig_5}
	\end{figure*}	

\subsection{Accuracy}

To compare their registration accuracy, the proposed method and other three methods are tested on six data sets, where the initial rigid transformations are estimated by the feature match method \cite{lei2017fast}. Experimental results are also reported in the form of registration errors. Table.~\ref{table_4} illustrates registration results of all compared methods tested on six data sets. For the facilitate comparison, Fig.~\ref{fig_5} also displays registration results in the form of cross-section. What's more, we also test these compared method on data sets contaminated by noises, where two levels of Gaussian noises are randomly added to all data points. Considering the randomness, each group of experiments are carried out 30 independently tests under each level of Gaussian noise. Accordingly, Table.~\ref{table_5} and Table.~\ref{table_6} display the statistics registration results for SNR=50dB and SNR=25dB Gaussian noises, respectively. As shown in Tables.~\ref{table_4}-\ref{table_6} and Fig.~\ref{fig_5}, except for the translation error of the Stanford Dragon, our method obtains the most accurate registration result for both noise-free data sets and noise contaminated data sets. Considering both rotation and translation errors, our method can also obtain the most accurate registration results for the Stanford Dragon.

	\begin{table*}[!t]
	    \centering
	    \caption{The registration errors on each data set using different method under random Gaussian noises with SNR=50dB (mean$\pm$std.), where the numbers in bold means the best performance.}
	    \begin{tabular}{c|c|c|c|c|c|c|c}
	    \toprule
	    Method & ~ & Angel (dm) & Armadillo (mm) & Bunny (mm) & Buddha (mm) & Dragon (mm) & Hand (dm) \\
	    \midrule    
	    \multirow{2}{*}{JRMPC} & ${{e}_{R}}$ & 0.0305 $\pm $ 0.0012 & 0.0171 $\pm $ 0.0003 & 0.0177 $\pm $ 0.0001 & 0.0375 $\pm $ 0.0013 & 0.0562 $\pm $ 0.0003 & 0.0078 $\pm $ 0.0002 \\
		& ${{e}_{t}}$ & 6.9544 $\pm $ 0.2390 & 1.3232 $\pm $ 0.0303 & 1.5743 $\pm $ 0.0168 & 1.8878 $\pm $ 0.0319 & 4.9795 $\pm $ 0.0213 & 0.8191 $\pm $ 0.0073 \\
		\midrule    
	    \multirow{2}{*}{TMM} & ${{e}_{R}}$ & 0.0086 $\pm$ 0.0004 & 0.0189 $\pm$ 0.0016 & 0.0085 $\pm$ 0.0004 & 0.0147 $\pm$ 0.0009 & 0.0259 $\pm$ 0.0007 & 0.0077 $\pm$ 0.0002 \\
		& ${{e}_{t}}$ & 2.8280 $\pm$ 0.0803 & 2.6052 $\pm$ 0.0613 & 0.7336 $\pm$ 0.0523 & 1.1703 $\pm$ 0.0464 & \bf{1.4834} $\pm$ \bf{0.0623} & 0.8306 $\pm$ 0.0068 \\
		\midrule    
	    \multirow{2}{*}{EMPMR} & ${{e}_{R}}$ & 0.0008 $\pm$ 0.0000 & 0.0206 $\pm$ 0.0000 & 0.0035 $\pm$ 0.0000 & 0.0083 $\pm$ 0.0000 & 0.0624 $\pm$ 0.0001 & 0.0012 $\pm$ 0.0000 \\
		& ${{e}_{t}}$ & 0.1762 $\pm$ 0.0034 & 1.7545 $\pm$ 0.0115 & 0.3444 $\pm$ 0.0024 & 1.3012 $\pm$ 0.0022 & 4.4119 $\pm$ 0.0069 & 0.0836 $\pm$ 0.0010 \\
		\midrule    
		\multirow{2}{*}{Our Method} & ${{e}_{R}}$ & \bf{0.0006} $\pm$ \bf{0.0000} & \bf{0.0054} $\pm$ \bf{0.0000} & \bf{0.0032} $\pm$ \bf{0.0000} & \bf{0.0065} $\pm$ \bf{0.0001} & \bf{0.0130} $\pm$ \bf{0.0000} & \bf{0.0006} $\pm$ \bf{0.0000} \\
		& ${{e}_{t}}$ & \bf{0.1072} $\pm$ \bf{0.0018} & \bf{0.7342} $\pm$ \bf{0.0018} & \bf{0.2594} $\pm$ \bf{0.0017} & \bf{1.1563} $\pm$ \bf{0.0098} & 1.5207 $\pm$ 0.0034 & \bf{0.0470} $\pm$ \bf{0.0008} \\
		\toprule
	    \end{tabular}
	    \label{table_5}
	\end{table*}
	
	\begin{table*}[!t]
	    \centering
	    \caption{The registration errors on each data set using different method under random Gaussian noises with SNR=25dB (mean$\pm$std.), where the numbers in bold means the best performance.}
	    \begin{tabular}{c|c|c|c|c|c|c|c}
	    \toprule
	    Method & ~ & Angel (dm) & Armadillo (mm) & Bunny (mm) & Buddha (mm) & Dragon (mm) & Hand (dm) \\
	    \midrule    
	    \multirow{2}{*}{JRMPC} & ${{e}_{R}}$ & 0.0296 $\pm $ 0.0025 & 0.0174 $\pm $ 0.0004 & 0.0178 $\pm $ 0.0008 & 0.0369 $\pm $ 0.0018 & 0.0567 $\pm $ 0.0008 & 0.0086 $\pm $ 0.0005 \\
		& ${{e}_{t}}$ & 6.6343 $\pm $ 0.4499 & 1.2604 $\pm $ 0.0627 & 1.5449 $\pm $ 0.0795 & 1.8373 $\pm $ 0.0788 & 5.0030 $\pm $ 0.0571 & 0.8365 $\pm $ 0.0154 \\
		\midrule    
	    \multirow{2}{*}{TMM} & ${{e}_{R}}$ & 0.0087 $\pm$ 0.0003 & 0.0188 $\pm$ 0.0015 & 0.0085 $\pm$ 0.0005 & 0.0143 $\pm$ 0.0008 & 0.0260 $\pm$ 0.0005 & 0.0077 $\pm$ 0.0003 \\
		& ${{e}_{t}}$ & 2.8735 $\pm$ 0.0784 & 2.5901 $\pm$ 0.0613 & 0.7493 $\pm$ 0.0656 & 1.1815 $\pm$ 0.05262 & \bf{1.4678} $\pm$ \bf{0.0501} & 0.8941 $\pm$ 0.0075 \\
		\midrule    
	    \multirow{2}{*}{EMPMR} & ${{e}_{R}}$ & 0.0008 $\pm$ 0.0001 & 0.0200 $\pm$ 0.0027 & 0.0035 $\pm$ 0.0002 & 0.0082 $\pm$ 0.0003 & 0.0616 $\pm$ 0.0052 & 0.0021 $\pm$ 0.0002 \\
		& ${{e}_{t}}$ & 0.1833 $\pm$ 0.0266 & 1.7252 $\pm$ 0.1930 & 0.3374 $\pm$ 0.0197 & 1.2745 $\pm$ 0.0352 & 4.4183 $\pm$ 0.1718 & 0.1467 $\pm$ 0.0111 \\
		\midrule    
    	\multirow{2}{*}{Our Method} & ${{e}_{R}}$ & \bf{0.0007} $\pm$ \bf{0.0001} & \bf{0.0054} $\pm$ \bf{0.0002} & \bf{0.0033} $\pm$ \bf{0.0002} & \bf{0.0066} $\pm$ \bf{0.0003} & \bf{0.0131} $\pm$ \bf{0.0003} & \bf{0.0020} $\pm$ \bf{0.0002} \\
		& ${{e}_{t}}$ & \bf{0.1407} $\pm$ \bf{0.0219} & \bf{0.7331} $\pm$ \bf{0.0135} & \bf{0.2758} $\pm$ \bf{0.0166} & \bf{1.1463} $\pm$ \bf{0.0389} & 1.5314 $\pm$ 0.0366 & \bf{0.1461} $\pm$ \bf{0.0119} \\
		\toprule
	    \end{tabular}
	    \label{table_6}
	\end{table*}

Actually, all compared methods utilizes the EM algorithm to optimize rigid transformations. The difference between them is the assumption, where each data point is sampled from different mixture probabilistic model. As mentioned before, JRMPC assumes that all point points are realizations of one central GMM and it should optimize massive Gaussian components, which make it easy to be trapped into local maximum. Besides, one weighted uniform distribution component is added into GMM, where the weight should be manually tuned to eliminate the influence of outliers. What's more, the GMM may unduly fit data noises with longer than Gaussian tail. All of these reasons will make JRMPC difficult to obtain accurate registration results. As displayed in Tables.~\ref{table_4}-\ref{table_6} and Fig.~\ref{fig_5}, JRMPC achieves the multi-view registration with the worst accuracy.

To account for outliers and noises with heavy-tail, TMM replace GMM by StMM to formulate multi-view registration problem. Since the t-distribution contains more heavy tail than that of Gaussian distribution, StMM can model scan data without any prior knowledge of the degree of outliers and noises. As shown in Tables.~\ref{table_4}-~\ref{table_6} and Fig.~\ref{fig_5}, TMM can obtain more accurate registration results than that of JRMPC. However, it also requires to optimize massive model parameters, which will inevitably reduce its registration accuracy. To address this problem, EMPMR supposes that each data point is
drawn from one unique GMM with equal covariance and membership probabilities. Since all Gaussian components can be determined by the efficient NN search method, it only requires to estimate $M$ rigid transformations as well as one Gaussian covariance. Accordingly, EMPMR is more likely to obtain the desired registration results. As shown in Tables.~\ref{table_4}-~\ref{table_6} and Fig.~\ref{fig_5}, EMPMR can obtain more accurate registration results than JRMPC and TMM except for Stanford Armadillo and Stanford Dragon. In Stanford Armadillo, there are many outliers. Besides, Stanford Buddha may contain noises with heavy-tail. Without any prior knowledge of noises and outliers, GMM may unduly fit these data points, which lead to undesired registration results for EMPMR.

While, the proposed method takes the advantages of TMM and EMPMR. It supposes that each data point is drawn from one unique StMM, where its NNs in other point sets are regarded as the t-distribution centroids with equal covariances, membership probabilities, and the fixed degrees of freedom. Subsequently, it only requires to optimize $M$ rigid transformation as well as one t-distribution covariance, which make it easy to obtain the desired registration results. What's more, the StMM can model noise contaminated data and outliers without any prior knowledge. As illustrated in Tables.~\ref{table_4}-~\ref{table_6} and Fig.~\ref{fig_5}, the proposed method is able to obtain the most accurate registration results for all six data sets.
	
\subsection{Efficiency}

To analyze the computation complexity, we restate that there are $M$ point sets being registered, where the $i$th point set contains $N_i$ data points and $M^{'}= M-1$. The number of required iterations is regarded as $K$ in the registration method. Before iteration, we build $k$-d tree for each point set to accelerate the NN search. For each point set, the complexity is $O(N_ilogN_i)$, which leads to the total complexity of $O(MN_ilogN_i)$ for $M$ point sets. In each iteration, the proposed method includes four operations. 
\begin{table*}[!t]
	    \centering
	    \caption{Comparison of computation complexity for different methods, where the value of $G$ is usually large than the value of $M$.}
	    \begin{tabular}{c|c|c|c|c}
	    \toprule
	     & JRMPC & TMM & EMPMR & Our method  \\
	    \midrule
	  Build $k$-d tree & -& -& $O(MN_ilog(N_i))$ & $O(MN_ilog(N_i))$     \\
	  Compute posterior probability & $O(KMGN_i)$ & $O(2KMGN_i)$    & $O(KMM^{'}N_i)$  & $O(2KMM^{'}N_i)$ \\
		  Update transformation & $O(KMGN_i)$  & $O(KMGN_i)$  &  $O(KMM^{'}N_i)$  & $O(KMM^{'}N_i)$  \\
		 Update cluster centroid & $O(KMGN_i)$  & $O(KMGN_i)$  & $O(KMN_ilog(N_i))$  & $O(KMN_ilog(N_i))$  \\
		 Update cluster covariance & $O(KMGN_i)$ &  $O(KMGN_i)$& $O(KMM^{'}N_i)$   & $O(KMM^{'}N_i)$  \\
	    \toprule
	    \end{tabular}
	    \label{table_7}
	\end{table*}
	
\begin{figure}[!t]
		\centering
		\includegraphics[width=0.9\columnwidth]{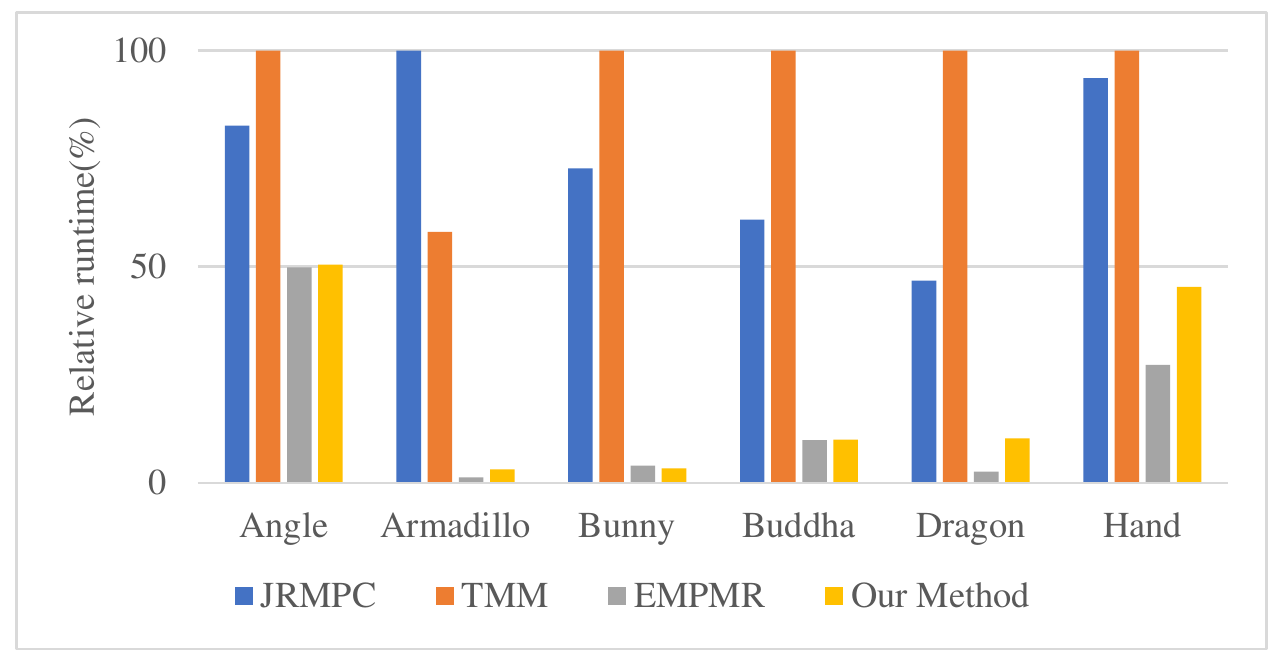}
		\caption{Comparison of averaging runtime over 10 independent tests on for different registration methods. For display convenience, relative run time is illustrated for each data set, where the most time-consuming method corresponds to 100$\%$ runtime.}
		\label{fig_6}
	\end{figure}

1) Compute posterior probability. For each data point $x_{i,l}$, there are two hidden variables $z_{i,l}$ and $u_{i,l}$, each of which can be assigned with $M^{'}$ values. Therefore, it requires the complexity of $O(2M^{'})$ to compute the posterior probabilities for the data point $x_{i,l}$. Since there are $N_i$ data points in the $i$th point set, the complexity is $O(2M^{'}N_i)$. Given $M$ point sets, the total computation complexity is $O(2KMM^{'}N_i)$ for $K$ iterations. 

2) Update the rigid transformation. To update the $i$th rigid transformation, $M^{'}$ point pairs are established for the data point $x_{i,l}$. As all data points in the $i$th point set are utilized to update the $i$th rigid transformation, the complexity is $O(M^{'}N_i)$. To update $M$ rigid transformations, the total computation complexity is $O(KMM^{'}N_i)$ for $K$ iterations.

3) Update the cluster centroid. For each data point, there are $M^{'}$ centroids in the corresponding StMM and they are determined by the NN search method with the complexity $O(M^{'}logN_i)$. As the $i$th contains $N_i$ data points, the complexity is $O(M^{'}N_ilogN_i)$. Given $M$ point sets, the total computation complexity is $O(KM^{'}N_ilogN_i)$ for $K$ iterations.

4) Update the cluster covariance. As all t-distribution components share the same covariance, there is only one covariance required to be updated. This covariance is updated by all established point pairs, where the number of point pairs is $O(MM^{'}N_i)$ for $M$ point sets. Therefore, the total computation complexity is $O(KMM^{'}N_i)$ for $K$ iterations.

Table \ref{table_7} lists the total computation complexity for each operation in the proposed method. For comparison, we also list the total computation complexity for each operation in other compared methods, where $G$ denotes the number of distribution components in JRMPC and TMM. Usually, $G$ is more large than $M^{'}$ and $logN_i$, so the proposed method is more efficient than JRMPC and TMM. But it is slightly less efficient than EMPMR due to more hidden variables required to be estimated.

Further, we tested the proposed method on these six data sets and compared it with other three methods. Specifically, each group of experiment was tested 10 times independently to eliminate randomness and the results are illustrated in Fig.~\ref{fig_6}. Obviously, the runtime of TMM and JRMPC are at the same level, which is much more than that of EMPMR and the proposed method. This is because each clustered centroid of TMM and JRMPC is estimated from all data points. It is required to build the correspondences between each cluster and all data points by the calculation of large weight matrix, which is really time consuming. While all clustered centroids of EMPMR and our method are determined by the NN search method, which can be efficiently solved by $k$-d tree based search method. What's more, both TMM and JRMPC should estimate a larger number of Gaussian variances. While, EMPMR and our method only require to estimate one variance. Therefore, they are more efficient than TMM and JRMPC. However, the formula of the t-distribution is a little more complicated than that of Gaussian distribution. Compared with GMM, StMM need to estimate an extra set of hidden variables, so EMPMR is a little more efficient than our method. Overall, the efficiency of our method is comparable with that of EMPMR. These conclusions are consistent with the theory analysis of computation complexity.

\section{Conclusions}
	This paper proposes an effective method for registration of multi-view point sets based on the StMM. It assumes that each data point is generated from one unique StMM, where its NNs in other point sets are regarded as the t-distribution centroids with equal covariances, membership probabilities, and the fixed degrees of freedom. Based on this assumption, the registration problem is formulated as the maximum likelihood estimation, which is reasonably optimized by the EM algorithm. Compared with most mixture probabilistic model methods, the proposed method only requires to estimate rigid transformations as well as one covariance. Therefore, it is more likely convergent to desired registration results. Since the StMM components is automatically determined by the number of point sets being registered, there is no trade-off between efficient and accuracy in the proposed method. What’s more, the proposed method takes the noise with heavy-tail into consideration by the t-distribution, so it is inherently robust to noises and outliers. Experimental results tested six bench mark datasets illustrate it outperforms state-of-the-art methods on robustness and accuracy.

	\section*{Acknowledgment}
This research was supported by the National Natural Science Foundation of China (No. 61573273), in part by the Fundamental Research Funds for Central Universities (No. xzy012019045).

	
	%
	

	\ifCLASSOPTIONcaptionsoff
	\newpage
	\fi

	
	
	\bibliographystyle{IEEEtran}
	\bibliography{IEEEfull}
	%

	%


	
	

\end{document}